\documentclass[10pt,conference,a4paper]{IEEEtran}
\ifCLASSINFOpdf
\else
\fi

\usepackage{graphicx}
\usepackage{url}
\usepackage{times}
\usepackage{latexsym}
\usepackage{graphics}
\usepackage{enumitem}
\usepackage{amssymb}
\usepackage{amstext}
\usepackage{lscape}
\usepackage{mathtools}
\usepackage[usenames,dvipsnames]{color}
\usepackage{soul}
\usepackage{epstopdf}
\usepackage{array}
\usepackage{adjustbox}
\makeatletter
\newcommand{\thickhline}{%
    \noalign {\ifnum 0=`}\fi \hrule height 1pt
    \futurelet \reserved@a \@xhline
}
\newcolumntype{"}{@{\hskip\tabcolsep\vrule width 1pt\hskip\tabcolsep}}
\makeatother

\newcommand{\DEL}[1]{}

\DeclarePairedDelimiter\floor{\lfloor}{\rfloor}

\newlist{inlinelist}{enumerate*}{1}
\setlist*[inlinelist,1]{%
  label=(\roman*),
}

\let\svthefootnote\thefootnote

\begin{document}
\setlength{\parindent}{0pt}

\title{Context Aware Nonnegative Matrix Factorization Clustering}
%

%
%

\author{\IEEEauthorblockN{Rocco Tripodi*}
\IEEEauthorblockA{ECLT, Ca' Foscari University\\ Ca' Minich 2940, Venice, Italy\\
Email: rocco.tripodi@unive.it}
\and

\IEEEauthorblockN{Sebastiano Vascon*}
\IEEEauthorblockA{ECLT, Ca' Foscari University\\ Ca' Minich 2940, Venice, Italy\\
Email: sebastiano.vascon@unive.it}
\and

\IEEEauthorblockN{Marcello Pelillo}
\IEEEauthorblockA{ECLT, Ca' Minich 2940, Venice, Italy \\ DAIS, Ca' Foscari University\\ Via Torino, Venice, Italy\\
Email: pelillo@unive.it}}
%

\maketitle

\begin{abstract}%
In this article we propose a method to refine the clustering results obtained
with the nonnegative matrix factorization (NMF) technique, imposing consistency
constraints on the final labeling of the data. The research community focused
its effort on the initialization and on the optimization part of this method,
without paying attention to the final cluster assignments. We propose a game
theoretic framework in which each object to be clustered is represented as a
player, which has to choose its cluster membership. The information obtained
with NMF is used to initialize the strategy space of the players and a weighted
graph is used to model the interactions among the players. These interactions
allow the players to choose a cluster which is coherent with the clusters chosen
by similar players, a property which is not guaranteed by NMF, since it produces
a soft clustering of the data. The results on common benchmarks show
that our model is able to improve the performances of many NMF formulations.%
 \end{abstract}
%
 
 %
\let\thefootnote\relax\footnotetext{* = equal contribution.}
\let\thefootnote\svthefootnote
 \section{Introduction}Nonnegative matrix factorization (NMF) is a particular
 kind of matrix decomposition in which an input matrix $X$ is factorized into
 two nonnegative matrices $W$ and $H$ of rank $k$, such that $WH^T$ approximates
 $X$. The significance of this technique is to find those vectors that are
 linearly independent in a determined vector space. In this way, they can be
 considered as the essential representation of the problem described by the
 vector space and can be considered as the latent structure of the data in a
 reduced space. The advantage of this technique, compared to other dimension
 reduction techniques such as Single Value Decomposition (SVD), is that the
 values taken by each vector are positive. In fact, this representation gives an
 immediate and intuitive glance of the importance of the dimensions of
 each vector, a characteristic that makes NMF particularly suitable for soft and
 hard clustering \cite{xu2003document}.\par The dimensions of $W$ and $H^T$ are
 $n \times k$ and $k \times m$, respectively, where $n$ is the number of
 objects, $m$ is the number of features, and $k$ is the number of dimensions of
 the new vector space. NMF uses different methods to initialize these matrices
 \cite{lee1999learning,wild2004improving,boutsidis2008svd} and then optimization
 techniques are employed to minimize the differences between $X$ and $WH^T$
 \cite{lin2007projected}.\par The initialization of the matrices $W$ and $H$
 \cite{boutsidis2008svd}, is crucial and can lead to different matrix
 decompositions, since it is performed randomly in many algorithms
 \cite{zhang2014nmf}. To the contrary, the step involving the final clustering
 assignment received less attention by the research community. In fact, once $W$
 and $H$ are computed, soft clustering approaches interpret each value in $W$ as
 the strength of association among objects and clusters and hard clustering
 approaches assign each object $j$ to the cluster $C_k$, where: \begin{equation}
 k = arg \textit{ } max (W_{j1},W_{j2},...,W_{jk}). \label{eq:argmax}
 \end{equation} \noindent %
 This step is also crucial since in hard clustering it could be the case that
 the assignments have to be made choosing among very similar (possibly equal)
 values and Equation \ref{eq:argmax} in this case can results inaccurate or even
 arbitrary. Furthermore this approach does not guarantee that the final
 clustering is consistent, with the drawback that very similar objects can
 result in different clusters. In fact, the clusters are assigned independently
 with this approach and two different runs of the algorithm can result in
 different partitioning of the data, due to the random initializations
 \cite{zhang2014nmf}. \par%
 These limitations can be overcome exploiting the relational information of the
 data and performing a consistent labeling\DEL{ of the data}. For this reason in this
 paper we use a powerful tool derived from evolutionary game theory, which
 allows to re-organize the clustering obtained with NMF, making it consistent
 with the structure of the data. With our approach we impose that the cluster
 membership has to be re-negotiated for all the objects. To this end, we employ
 a dynamical system perspective, in which it is imposed that similar objects
 have to belong to similar clusters, so that the final clustering will be
 consistent with the structure of the data. This perspective has demonstrated its efficacy in different semantic categorization scenarios \cite{tripodicl2016, 2016arXiv160702436T}, which involve a high number of interrelated categories and require the use of contextual and similarity information.
\section{NMF Clustering}%
 NMF is employed as clustering algorithm in different applications. It has been
 successfully applied in “parts-of-whole” decomposition \cite{lee1999learning},
 object clustering \cite{ding2006nonnegative}, face recognition
 \cite{wang2005non}, multimedia analysis \cite{caicedo2012multimodal}, and DNA
 gene expression grouping \cite{zhang2010binary}. It is an appealing method
 because it can be used to perform together objects and feature clustering. The
 generation of the factorized matrices starts from the assumption that the
 objects of a given dataset belong to $k$ clusters and that these clusters can
 be represented by the features of the matrix $W$, which denotes the relevance
 that each cluster has for each object. This description is very useful in soft
 clustering applications because an object can contain information about
 different clusters in different measure. For example a text about a the launch
 of a new car model into the marked can contain information about economy,
 automotive or life-style, in different proportions. Hard clustering
 applications require to choose just one of these topics to partition the data
 and this can be done considering not only the information about the single
 text, but also the information of the other texts in the texts collection, in
 order to divide the data in coherent groups. \par In many algorithms the
 initialization of the matrices $W$ and $H$ is done randomly
 \cite{lee1999learning} and have the drawback to always lead to different
 clustering results. In fact, NMF converges to local minima and for this reason
 has to be run several times in order to select the solution that approximates
 better the initial matrix. To overcome this limitation there were proposed
 different approaches to find the best initializations based on feature
 clustering \cite{wild2004improving} and SVD techniques \cite{boutsidis2008svd}.
 These initializations allow NMF to converge always to the same solution.
 \cite{wild2004improving} uses spherical $k$-means to partition the columns of
 $X$ into $k$ clusters and selects the centroid of each cluster to initialize
 the corresponding column of $W$. Nonnegative Double Singular Value
 Decomposition (NNDSVD) \cite{boutsidis2008svd} computes the $k$ singular
 triplets of $X$, forms the unit rank matrices using the singular vector pairs,
 extracts from them their positive section and singular triplets and with this
 information initializes $W$ and $H$. This approach has been shown to be almost
 as good as that obtained with random initialization
 \cite{boutsidis2008svd}.\par A different formulation of NMF as clustering
 algorithm was proposed by \cite{kuang2015symnmf} (SymNMF). The main difference
 with classical NMF approaches is that SymNMF takes a square nonnegative
 similarity matrix as input instead of a $n \times m$ data matrix. It starts
 from the assumption that NMF was conceived as a dimension reduction technique
 and that this task is different from clustering. In fact, dimension reduction
 aims at finding a few basis vectors that approximate the data matrix and
 clustering aims at partitioning the data points where similarity is high among
 the elements of a cluster and low among the elements of different clusters.
 In this
 formulation a basis vector strictly represents a cluster. \par Common
 approaches obtain an approximation of $X$ minimizing the Frobenius norm of the
 difference $||X-WH^T||$ or the generalized Kullback-Leibler divergence
 $D_{KL}(X||WH^T)$ \cite{berman1979nonnegative} , using multiplicative update
 rules \cite{lee2001algorithms} or gradient methods \cite{lin2007projected}.

\section{Game Theory and Game Dynamics} \label{sec:GT}%
Game theory was introduced by Von Neumann and Morgenstern \cite{von1944theory}
in order to develop a mathematical framework able to model the essentials of
decision making in interactive situations. In its \textit{normal-form}
representation, it consists of a finite set of players $I=\{1,..,n\}$, a set of
pure strategies for each player $S_i=\{s_1, ..., s_n\}$, and a utility function
$u : S_1 \times ... \times S_n \rightarrow \mathbb{R}$, which associates
strategies to payoffs. Each player can adopt a strategy in order to play a game
and the utility function depends on the combination of strategies played at the
same time by the players involved in the game, not just on the strategy chosen
by a single player. An important assumption in game theory is that the players
are rational and try to maximize the value of $u$. Furthermore, in
\emph{non-cooperative games} the players choose their strategies independently,
considering what other players can play and try to find the best strategy
profile to employ in a game.\par Nash equilibria represent the key concept of
game theory and can be defined as those strategy profiles in which each strategy
is a best response to the strategy of the co-player and no player has the
incentive to unilaterally deviate from his decision, because there is no way to
do better. The players can also play \emph{mixed strategies}, which are
probability distributions over pure strategies. A mixed strategy profile can be
defined as a vector $x=(x_{1},\ldots,x_{m} )$, where $m$ is the number of pure
strategies and each component $x_{h}$ denotes the probability that the player
chooses its $h$th pure strategy. Each mixed strategy corresponds to a point on
the simplex and its corners correspond to pure strategies.\par In a
\emph{two-player game}, a strategy profile can be defined as a pair $(p,q)$
where $p \in \Delta_i$ and $q \in \Delta_j$. The expected payoff for this
strategy profile is computed as: \begin{equation}\label{eq:gtpayoff} u_i(p,q)=p
\cdot A_i q \text{ , } u_j(p,q)=q \cdot A_j p \end{equation} \noindent where
$A_i$ and $A_j$ are the payoff matrices of player $i$ and $j$ respectively.\par
In evolutionary game theory we have a population of agents which play games
repeatedly with their neighbors and update their beliefs on the state of the
system choosing their strategy according to what has been effective and what has
not in previous games, until the system converges. The strategy space of each
player $i$ is defined as a mixed strategy profile $x_i$, as defined above. The
payoff corresponding to a single strategy can be computed as:
\begin{equation}\label{eq:singlePayoff} u_i(e_i^h) = \sum_{j=1}^n(A_{ij} x_j)_h
\end{equation} \noindent and the average payoff is:
\begin{equation}\label{eq:averagePayoff} u_i(x) =\sum_{j=1}^n x_i^T A_{i j}x_j
\end{equation} \noindent where $n$ is the number of players with whom the games
are played and $A_{ij}$ is the payoff matrix among player $i$ and $j$. The
replicator dynamic equation \cite{taylor1978evolutionary} is used in order to
find those states, which correspond to the Nash equilibria of the games,
\begin{equation}\label{eq:replicatorDS} x^h(t+1)=x^h(t)\frac{u(e^h,x)}{u(x,x)}
\text{ } \forall h \in S \end{equation} \noindent This equation allows better
than average strategies to grow at each iteration and we can consider each
iteration of the dynamics as an \emph{inductive learning} process, in which the
players learn from the others how to play their best strategy in a determined
context.

 \IEEEpeerreviewmaketitle \begin{figure*} \centering
 \includegraphics[width=0.91\textwidth, trim={0cm 9.8cm 0cm 0.9cm},clip]{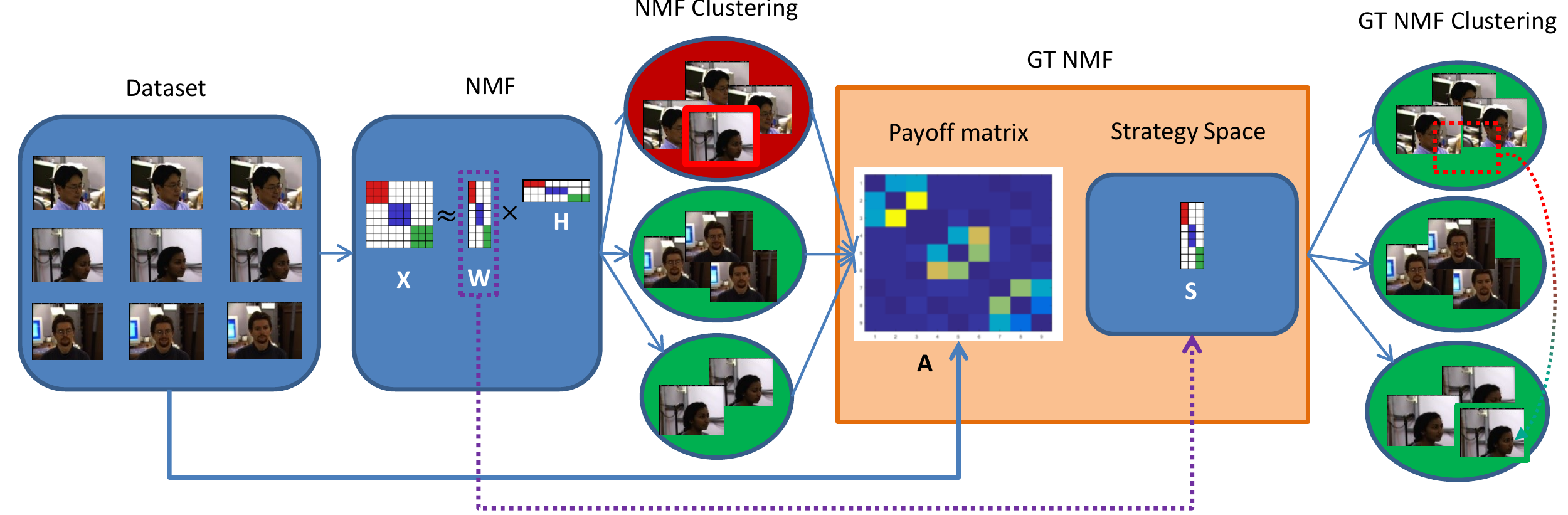} 
 \caption{The pipeline of the proposed game-theoretic refiner method: a dataset
is clustered using NMF obtaining a partition of the original data into $k$
clusters. A pairwise similarity matrix $A$ is constructed on the original set of
data and the clustering assignments obtained with NMF. The output of NMF ($W$) and
the matrix $A$ are used to refine the assignments. The matrix $W$ is also used
to initialize the strategy space of the games. In red the wrong assignment that
is corrected after the refinement. Best viewed in color.}\label{fig:pipeline} \end{figure*}%

 \section{Our Approach} In this section we present the Game Theoretic Nonnegative
Matrix Factorization (GTNMF), our approach to NMF clustering refinement. The
pipeline of this method is depicted in Fig. \ref{fig:pipeline}. We extract
the feature vectors of each object in a dataset then, depending on the NMF
algorithm used, we give as input to NMF the feature vectors or a similarity
matrix. GTNMF takes as input the matrix $W$ obtained with NMF and the similarity
graph $A$ (see Section \ref{sec:gamesGraph}) of the dataset to produce a
consistent clustering of the data. \par Each data point, in our formulation, is
represented as a player that has to choose its cluster membership. The weighted
graph $A$ measures the influence that each player has on the others. The matrix
$W$ is used to initialize the strategy space $S$ of the players. We use the
following equation $s_{ij}=\frac{w_{ij}}{\sum_{j=1}^K w_{ij}}$ to constrain the
strategy space of each player to lie on the standard simplex, as required in a
game theoretic framework (see Section \ref{sec:GT}). The dynamics 
are not started on the center of the $K$-dimensional simplex, as it is commonly
done in unsupervised learning tasks, but on a different interior point, which
corresponds to the solution point of NMF and do not compromise the dynamics to
converge to Nash equilibria \cite{weibull1997evolutionary}.\par Now that we
have the topology of the data $A$ and the strategy space of the game $S$ we can
compute the Nash equilibria of the games according to equation
(\ref{eq:replicatorDS}). In each iteration of the system each player plays a
game with its neighbors $N_i$ according to the similarity graph $A$ and the
payoffs are calculated as follows: \begin{equation}\label{eq:payoff}
u_i(e^h,s)=\sum_{j \in N_i}(a_{ij}s_j)_h 
\end{equation}
\noindent
and
\begin{equation}
u_i(s) =\sum_{j \in N_i} x_i^T (a_{ij}s_j)
\end{equation}
\noindent
 We assume that the payoff of player $i$ depends on the similarity that it has
 with player $j$, $a_{ij}$, and its preferences, ($s_j$). During each phase of
 the dynamics a process of selection allows strategies with higher payoff to
 emerge and at the end of the process each player chooses its cluster according
 to these constraints. Since Equation \ref{eq:replicatorDS} models a dynamical
 system it requires some criteria to stop. In the experimental part of this work
 we used as stopping criteria the maximum number of iterations $=100$ and
 $\delta<10^4$, where $\delta$ is the Euclidean norm between the strategy space
 at time $t$ and at time $t+1$. 
 \section{Experimental Setup and Results} In this section, we show the
 performances of GTNMF on different text and image datasets, and compare it with
 standard NMF\footnote{Code: \url{https://github.com/kimjingu/nonnegfac-matlab}}
 \cite{kim2014algorithms}, NMF-S \cite{kim2014algorithms} (same as NMF but with
 the similarity matrix as input instead of the features), SymNMF
 \cite{kuang2015symnmf}\footnote{Code:
 \url{https://github.com/andybaoxv/symnmf}} and NNDSVD\footnote{Code:
 \url{http://www.boutsidis.org/NNDSVD_matlab_implementation.rar}}
 \cite{boutsidis2008svd}, which use the standard maximization technique to
 obtain an hard clustering of the data.
 In Table \ref{tab:results} we refer to our approach
as \emph{NMF-algorithm+GT} which means that the GTNMF has been initializied with the particular NMF-algorithm.
\subsection{Datasets description} \begin{table*}[] \centering \caption{Datasets description. \textsuperscript{$\bigstar$} = the dataset has been pruned as in \protect\cite{kuang2015symnmf}.} \label{tbl:datasets} \begin{tabular}{|l|c|c|c|c|c|c|c|c|} \hline Dataset & Type & Points & Features & Clusters & Balance & Mean Clust & Min Cl Size & Max Cl Size \\ \hline
NIPS\textsuperscript{$\bigstar$} & text & 424 & 17522 & 9 & 0.105 & 47.1 & 15 & 143 \\ \hline
NIPS & text & 451 & 17522 & 13 & 0.028 & 34.7 & 4 & 143 \\ \hline
Reuters\textsuperscript{$\bigstar$} & text & 8095 & 14143 & 20 & 0.011 & 404.8 & 42 & 3735 \\ \hline
Reuters & text & 8654 & 14333 & 65 & 0 & 133.1 & 1 & 3735 \\ \hline
RCV1\textsuperscript{$\bigstar$} & text & 13254 & 20478 & 40 & 0.028 & 331.4 & 45 & 1587 \\ \hline
RCV1 & text & 13732 & 20478 & 75 & 0.001 & 183.1 & 1 & 1587 \\ \hline
PIE-Expr & image & 232 & 4096 & 68 & 0.75 & 3.4 & 3 & 4 \\ \hline
ORL & image & 400 & 5796 & 40 & 1 & 10 & 10 & 10 \\ \hline
COIL-20 & image & 1440 & 4096 & 20 & 1 & 72 & 72 & 72 \\ \hline
ExtYaleB & image & 2447 & 3584 & 38 & 0.678 & 64.4 & 59 & 87 \\ \hline
\end{tabular}\end{table*}
 The evaluation of GTNMF
has been conducted on datasets with different characteristics (see Table
\ref{tbl:datasets}). We used textual (Reuters, RCV1, NIPS) and image (COIL-20,
ORL, Extended YaleB and PIE-Expr) datasets. Authors in \cite{kuang2015symnmf}
discarded the objects belonging to small clusters in order to make the dataset
more balanced, simplifying the task. We tested our method using this approach
and also keeping the datasets as they are (without reduction), which lead to
situations in which it is possible to have in the same dataset clusters with
thousands of objects and clusters with just one object (e.g. RCV1). 
%
%
%
%
%
%
%
%
%

\subsection{Data preparation} \label{sec:dataPrep} The datasets have been
processed as suggested in \cite{kuang2015symnmf}. Given an $n \times m$ data
matrix $X$, the similarity matrix $A$ is constructed according to the type of
dataset (textual and image). With textual dataset each feature vector is
normalized to have unit 2-norm and the cosine distance is computed,
$A={x_i}^Tx_j$. For image datasets each feature (column) is first normalized to
lie in the range $[0,1]$ and then it is applied the following kernel:
$A_{i,j}=\exp\{- \frac{||x_i-x_j||}{\sigma_i\sigma_j}\}$, where $\sigma_i$ is
the Euclidean distance of the 7-th nearest neighbor \cite{zelnik2004self}. In
all cases $A_{ii}=0$.\par The matrix is thus sparsified keeping only the $q$
nearest neighbors for each point. The parameter $q$ is set accordingly to
\cite{von2007tutorial} and represents a theoretical bound that guarantees the
connectedness of a graph: \begin{equation}\label{eq:q} q=\floor{log_2(n)}+1
\end{equation} Let $N(i)=\{ j \mbox{ s.t. } x_j \in q\mbox{-NN of } i \}$ then
$A_{ij}= A_{ij}$ if $i \in N(j) \textit{ or } j \in N(i)$ and $0$ otherwise. 
%
The matrix $A$ is thus normalized in a normalized-cut fashion obtaining the
final matrix $A_{ij}=A_{ij}\sqrt{d_i}\sqrt{d_j}$ where
$d_i=\sum_{s=1}^n{A_{is}}$. The matrix $A$ is given as input to all the compared
methods, expect from NMF to which the data matrix $X$ is given. See
\cite{kuang2015symnmf} for further details on this phase.


\subsection{Games graph}\label{sec:gamesGraph} In Sec.\ref{sec:dataPrep} has been
explained how to create the similarity matrix for NMF, the same methodology has
been used to create the payoff matrix $A$ for the GTNMF, with the only
difference that, in this case, we exploit the partitioning obtained with NMF in
order to identify what could be the expected size of the clusters. The
assumption here is that the clustering obtained via NMF provides a good insight
on the size of the final clusters and accordingly with this information a proper
number $q$ (see Equation \ref{eq:q}) can be selected. A cluster $C$ can be
considered as a fully connected subgraph and thus the number of neighbors of
each element in the cluster $C$ should be at least $q_C=\floor{log_2(|C|)}+1$ to
guarantee the connectedness of the cluster itself. The variable $q$ is thus
chosen based on the same principle of \cite{von2007tutorial} but instead of
taking into account the entire set of points (as in Sec.\ref{sec:dataPrep}) we
focused only on the subsets induced by the NMF clustering. This results in
having a different $q$ for each point in the dataset based on the following
rule: \begin{equation} q_i= \floor{log_2(|C|)}+1 \label{eqn:q} \end{equation}
where $|C|$ is the cardinality of cluster $C$ to which the $i$-th element
belongs to. For obvious reason $q_i \leq q \mbox{ , } \forall i=1,\dots,n$ and
thus concentrating only on the potential number of neighbors that belong to the
cluster and not in the entire graph because we are doing a refinement. From a
game-theoretic perspective this means to focus the games only among a set of
similar players which are likely to belong to the same cluster.

\begin{table*}[ht] \centering 
 \caption{Performance of GTNMF compared to several NMF approaches. The mean and std deviation of 20 runs are reported.} \label{tab:results}
 \begin{adjustbox}{max width=\textwidth}
\begin{tabular}{|l"l|l"l|l"l|l"l|l|} \hline
 \multicolumn{9}{c}{\textbf{Normalized Mutual Information}} \\ \hline
Dataset & SymNMF & SymNMF+GT & NMF & NMF+GT &  NMFS &  NMFS+GT & NNDSVD & NNDSVD+GT \\ \hline
NIPS \textsuperscript{$\bigstar$} & 0.385 ($\pm 0.011$) & \textbf{0.405} ($\pm 0.016$) & 0.375 ($\pm 0.022$) & \textbf{0.386} ($\pm 0.011$) & 0.388 ($\pm 0.006$) & \textbf{0.403} ($\pm 0.007$) & 0.388 & \textbf{0.399}   \\ \hline
NIPS & 0.387 ($\pm 0.007$) & \textbf{0.418} ($\pm 0.017$) & 0.401 ($\pm 0.016$) & \textbf{0.406} ($\pm 0.016$) & 0.393 ($\pm 0.008$) & \textbf{0.412} ($\pm 0.018$) & 0.388 & \textbf{0.421}   \\ \hline
Reuters \textsuperscript{$\bigstar$} & 0.502 ($\pm 0.014$) & \textbf{0.51} ($\pm 0.016$) & 0.451 ($\pm 0.026$) & \textbf{0.49} ($\pm 0.02$) & 0.505 ($\pm 0.014$) & \textbf{0.511} ($\pm 0.015$) & \textbf{0.427} & 0.425 \\ \hline
Reuters & 0.517 ($\pm 0.007$) & \textbf{0.525} ($\pm 0.006$) & 0.442 ($\pm 0.006$) & \textbf{0.497} ($\pm 0.003$) & 0.518 ($\pm 0.006$) & \textbf{0.527} ($\pm 0.005$) & 0.488 & \textbf{0.493}   \\ \hline
RCV1 \textsuperscript{$\bigstar$} & 0.406 ($\pm 0.007$) & \textbf{0.422} ($\pm 0.007$) & 0.51 ($\pm 0.007$) & \textbf{0.516} ($\pm 0.005$) & 0.404 ($\pm 0.009$) & \textbf{0.42} ($\pm 0.01$) & \textbf{0.403} & 0.402 \\ \hline
RCV1 & 0.411 ($\pm 0.006$) & \textbf{0.422} ($\pm 0.006$) & 0.462 ($\pm 0.009$) & \textbf{0.483} ($\pm 0.007$) & 0.413 ($\pm 0.006$) & \textbf{0.424} ($\pm 0.006$) & 0.398 & \textbf{0.407}   \\ \hline
PIE-Expr & 0.95 ($\pm 0.004$) & \textbf{0.968} ($\pm 0.004$) & 0.939 ($\pm 0.008$) & \textbf{0.959} ($\pm 0.006$) & 0.89 ($\pm 0.005$) & \textbf{0.931} ($\pm 0.006$) & 0.86 & \textbf{0.889}   \\ \hline
ORL & 0.888 ($\pm 0.006$) & \textbf{0.921} ($\pm 0.006$) & 0.691 ($\pm 0.015$) & \textbf{0.844} ($\pm 0.014$) & 0.889 ($\pm 0.006$) & \textbf{0.918} ($\pm 0.004$) & 0.808 & \textbf{0.892}   \\ \hline
COIL-20 & 0.871 ($\pm 0.009$) & \textbf{0.875} ($\pm 0.012$) & 0.619 ($\pm 0.017$) & \textbf{0.669} ($\pm 0.016$) & 0.877 ($\pm 0.013$) & \textbf{0.883} ($\pm 0.013$) & 0.824 & \textbf{0.836}   \\ \hline
ExtYaleB & 0.308 ($\pm 0.005$) & \textbf{0.313} ($\pm 0.005$) & \textbf{0.356} ($\pm 0.006$) & 0.355($\pm 0.007$) & 0.309 ($\pm 0.007$) & \textbf{0.314} ($\pm 0.005$) & 0.288 & \textbf{0.315}   \\ \hline
 \multicolumn{9}{c}{\textbf{Accuracy}} \\ \hline
Dataset & SymNMF & SymNMF+GT & NMF & NMF+GT &  NMFS &  NMFS+GT & NNDSVD & NNDSVD+GT \\ \hline
NIPS \textsuperscript{$\bigstar$} & 0.462 ($\pm 0.013$) & \textbf{0.483} ($\pm 0.016$) & \textbf{0.426} ($\pm 0.02$) & 0.425($\pm 0.014$) & 0.465 ($\pm 0.011$) & \textbf{0.485} ($\pm 0.017$) & 0.474 & \textbf{0.509}   \\ \hline
NIPS & 0.379 ($\pm 0.01$) & \textbf{0.415} ($\pm 0.037$) & \textbf{0.396} ($\pm 0.022$) & 0.39($\pm 0.018$) & 0.384 ($\pm 0.014$) & \textbf{0.407} ($\pm 0.031$) & 0.466 & \textbf{0.503}   \\ \hline
Reuters \textsuperscript{$\bigstar$} & 0.517 ($\pm 0.044$) & \textbf{0.528} ($\pm 0.043$) & 0.322 ($\pm 0.024$) & \textbf{0.401} ($\pm 0.026$) & 0.516 ($\pm 0.037$) & \textbf{0.525} ($\pm 0.037$) & 0.403 & \textbf{0.427}   \\ \hline
Reuters & 0.324 ($\pm 0.029$) & \textbf{0.363} ($\pm 0.032$) & 0.222 ($\pm 0.011$) & \textbf{0.282} ($\pm 0.02$) & 0.339 ($\pm 0.023$) & \textbf{0.378} ($\pm 0.024$) & 0.277 & \textbf{0.339}   \\ \hline
RCV1 \textsuperscript{$\bigstar$} & \textbf{0.292} ($\pm 0.015$) & 0.289($\pm 0.014$) & 0.383 ($\pm 0.009$) & \textbf{0.387} ($\pm 0.01$) & \textbf{0.298} ($\pm 0.017$) & 0.297($\pm 0.017$) & \textbf{0.285} & 0.276 \\ \hline
RCV1 & 0.243 ($\pm 0.008$) & \textbf{0.247} ($\pm 0.008$) & 0.279 ($\pm 0.01$) & \textbf{0.295} ($\pm 0.011$) & 0.242 ($\pm 0.01$) & \textbf{0.245} ($\pm 0.011$) & 0.239 & \textbf{0.24}   \\ \hline
PIE-Expr & 0.81 ($\pm 0.021$) & \textbf{0.85} ($\pm 0.019$) & 0.783 ($\pm 0.023$) & \textbf{0.809} ($\pm 0.024$) & 0.617 ($\pm 0.019$) & \textbf{0.7} ($\pm 0.02$) & \textbf{0.536} & 0.513 \\ \hline
ORL & 0.776 ($\pm 0.017$) & \textbf{0.811} ($\pm 0.018$) & 0.465 ($\pm 0.019$) & \textbf{0.608} ($\pm 0.026$) & 0.77 ($\pm 0.013$) & \textbf{0.804} ($\pm 0.015$) & 0.653 & \textbf{0.71}   \\ \hline
COIL-20 & 0.727 ($\pm 0.036$) & \textbf{0.729} ($\pm 0.037$) & 0.478 ($\pm 0.023$) & \textbf{0.507} ($\pm 0.025$) & 0.739 ($\pm 0.046$) & \textbf{0.741} ($\pm 0.046$) & \textbf{0.674} & 0.672 \\ \hline
ExtYaleB & \textbf{0.235} ($\pm 0.008$) & 0.228($\pm 0.007$) & 0.194 ($\pm 0.007$) & \textbf{0.197} ($\pm 0.009$) & \textbf{0.237} ($\pm 0.012$) & 0.23($\pm 0.01$) & 0.229 & \textbf{0.242}   \\ \hline
 \end{tabular} \end{adjustbox}
 \end{table*}

\subsection{Evaluation measures} \DEL{The evaluation of o}Our approach has been
validated using two different measures, accuracy (AC) and normalized mutual
information (NMI). AC is calculated as
$\frac{\sum_{i=1}^n{\delta(\alpha_i,map(l_i))}}{n}$, where $n$ denotes the total
number of documents in the dataset, $\delta(x,y)$ equals to 1 if $x$ and $y$ are
clustered in the same class; $map(L_i)$ maps each cluster label $l_i$ to the
equivalent label in the benchmark. The best mapping is computed using the
Kuhn-Munkres algorithm \cite{lovasz1986matching}. The AC counts the number
of correct clusters assignments. NMI indicates the level of agreement between
the clustering $C$ provided by the ground truth and the clustering $C'$ produced
by a clustering algorithm. The mutual information (MI) between the two
clusterings is computed as, \begin{equation}\label{eq:mi} \sum_{c_i \in C, c_j'
\in C'}p(c_i,c_j') \cdot log_2 \frac{p(c_i,c_j')}{p(c_i) \cdot p(c_j')}
\end{equation} \noindent where $p(c_i)$ and $p(c_i')$ are the probabilities that
a document of the corpus belongs to cluster $c_i$ and $c_i'$, respectively, and
$p(c_i,c_i')$ is the probability that the selected document belongs to $c_i$ as
well as $c_i'$ at the same time. The MI information is then normalized with the
following equation, \begin{equation}\label{eq:nmi}
NMI(C,C')=\frac{MI(C,C')}{max(H(C),H(C'))} \end{equation} \noindent where $H(C)$
and $H(C')$ are the entropies of $C$ and $C'$, respectively.

























 
\subsection{Evaluation} The results of our evaluation are shown in Table
\ref{tab:results}, where we reported the mean and standard deviation of 20
independent runs. For NNDSVD the experiments are run only one time, since it
converges always to the same solution. 
The performances of GTNMF in most of the
cases are higher those of the different NMF algorithms. In particular, we can
notice that despite the different settings (textual/image datasets) our
algorithm is able improve the NMI performance in 33/36 cases with a maximum gain
of $\simeq 15.3\%$ (which is quite impressive) and a maximum loss of $0.2\%$.%
\begin{figure}[h!] \centering \includegraphics[width=0.49\textwidth, trim={0.5cm
5.8cm 0cm 0cm},clip]{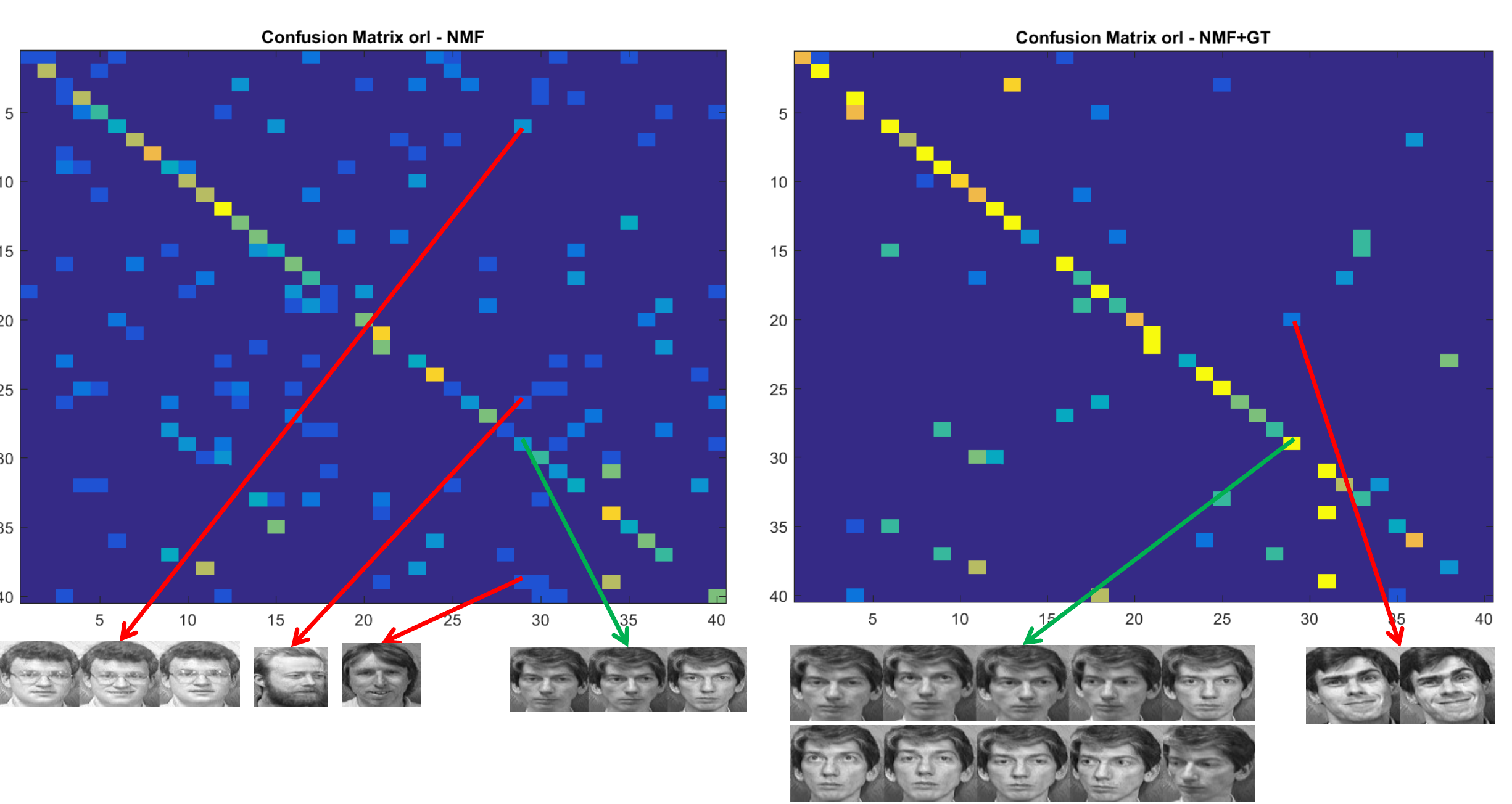} \caption{On the left side a
confusion matrix produced by NMF on the ORL dataset and on the right side the
ones produced by our method.}\label{fig:rightCaseStudy} \end{figure} 
constant gain in the NMI means, in practice, that the algorithm is able to
partition better the dataset, making the final clustering closer to the ground
truth. In terms of AC, on $27/36$ cases the method improve on the compared
methods, with a maximum gain of $14.3\%$ and maximum loss of $2.3\%$. It worth
noting that the negative results are very low and in most of the cases the
corresponding number of incorrect reallocations is low, in fact, $-0.6\%$ in the
NIPS dataset means $2.7$ elements or $-0.2\%$ in COIL-20 corresponds to $2.8$
elements.\par
\begin{figure*}[ht!] \centering \includegraphics[width=1\textwidth]{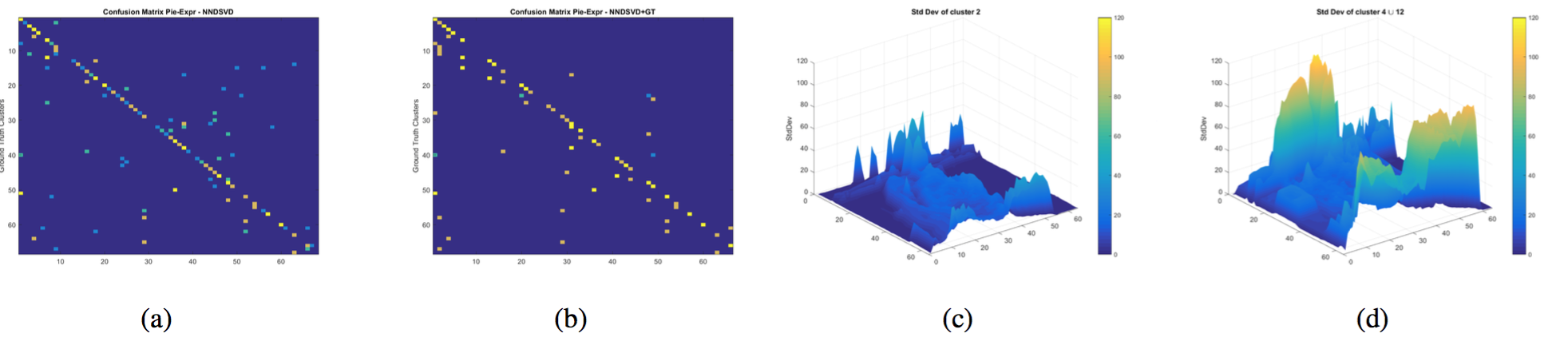}
 \caption{On a and b the confusion matrices produced by NMF and GTNMF on Pie-Expr. On c the std dev of the objects merged together by GTNMF and on d the std dev of two
random clusters combined together.}\label{fig:wrongCaseStudy}
\end{figure*}

 The mean gain for NMI and AC are $2.68\%$ and $2.30\%$,
respectively, while the mean loss are $0.16\%$ and $0.01\%$. In some cases we
can see that we obtain a loss in NMI and a gain in AC, for example on ExtYaleB
with NMF. In this case the similarity matrix given as input to GTNMF tends to
concentrate more objects in the same cluster, because the dataset is not
balanced and it could be the case that, in these situations, a big cluster tends
to attract many objects, increasing the probability of good reallocations, which
results in an increase in AC and in a potentially wrong partitioning of the
data. To the contrary in some experiments we have a loss in AC and a gain in
NMI. For example on PIE-Expr we noticed that we are able to put together many
objects that the other approaches tend to keep separated, but in this particular
case GTNMF collected in the same cluster all the objects belonging to four
similar clusters and for this reason there was a loss in accuracy (see Fig.
\ref{fig:wrongCaseStudy}). \par We can see that the results of our method on
well balanced datasets (ORL, COIL-20) are almost always good. Also on very
unbalanced datasets, such as Reuters and Reuters\textsuperscript{$\bigstar$} we
have always good performances, whatever is the method used. These datasets
depict better real life situations and the improvements over them are due to the
fact that in these cases it is necessary to exploit the geometry of the data in
order to obtain a good partitioning.

%
%
%
%

A positive and a negative case study are shown in Fig.
\ref{fig:rightCaseStudy} and \ref{fig:wrongCaseStudy}, respectively. In Fig.
\ref{fig:rightCaseStudy} the confusion matrix obtained with GTNMF is less sparse
and more concentrated on the main diagonal. Given the same cluster Id, the NMF
method agglomerates different clusters (red arrows) while, after the refinement,
the number of elements corresponding to the correct cluster are moved. In Fig.
b the algorithm tends to agglomerates the elements on a
single cluster (second column of the matrix). This can be explained on how the
similarity matrix is composed and on the nature of the data: in Fig.
c the std dev of the images in the agglomerated cluster is
reported, as one can notice the std is very low meaning that all the faces in
that cluster are very similar to each other. To give a counterexample we report
on Fig. d the std dev of two random cluster joined together, is
straightforward to notice that the std dev is higher than in the previous
example meaning that the elements within those two clusters are highly
dissimilar in nature and thus easily separable.%

\section{Conclusion} In this work we presented GTNMF, a game theoretic model to
improve the clustering results obtained with NMF going beyond the classical
technique used to make the final clustering assignments. The $W$ matrix obtained
with NMF can have an high entropy which make the choice of a cluster very
difficult in many cases. With our approach we try to reduce the uncertainty in
the matrix $W$ using evolutionary dynamics and taking into account contextual
information to perform a consistent labeling of the data. In fact, with our
method similar objects are assigned to similar clusters, taking into account the
initial solution obtained with NMF.\par We conducted an extensive analysis of
the performances of our method and compared it with different NMF formulations
and on datasets with different features and of different kind. The results of
the evaluation demonstrated that our approach is almost always able to improve
the results of NMF and that when it have negative results those results are
practically non significant. The algorithm is quite general thanks to the
adaptive auto-tuning of the payoff matrix and can deal with balanced and
completely unbalanced datasets. \par As future work we are planning to use
different initialization of the strategy space, to use new similarity functions
to construct the games graph, to apply this method to different problems and to
different clustering algorithms. 
\section*{Acknowledgment}
This work was supported by Samsung Global Research Outreach Program.

%
%
%
%
%
%
%
%
%
%
%
%
%
%
%
%
%
%
%
%
%
%
%
%
%
%
%
%

\bibliographystyle{IEEEtran}
\bibliography{IEEEabrv,root}

\end{document}